\documentclass[runningheads]{llncs}

\usepackage{graphicx}
\usepackage{xcolor}
\usepackage{amsmath}
\usepackage{amssymb}
\usepackage{caption}
\usepackage{subcaption}
\usepackage{times}
\usepackage{hyperref}
\usepackage{wrapfig}

\newcommand{\myparagraph}[1]{\vspace{0.3cm}\noindent\textbf{#1}~}

\begin{document}

\title{Learning Section Weights for Multi-Label Document Classification}

\authorrunning{M. Moradi Fard et al.}

\author{Maziar Moradi Fard\inst{1} 
Paula Sorrolla Bayod\inst{1}\and
Kiomars Motarjem\inst{2}\and
Mohammad Alian Nejadi\inst{3}\and
Saber Akhondi\inst{1}\and
Camilo Thorne\inst{4}
}
\institute{Elsevier B.V., Radarweg 29a, 1043 NX Amsterdam, Netherlands \\ 
\email{m.moradifard@elsevier.com, s.akhondi@elsevier.com, p.sorollabayod@elsevier.com} \and
Depart of statistics Tarbiat Modares University, Jalal Al Ahmad Street 7, Tehran, Iran\\
\email{k.motarjem@modares.ac.ir} \and
University of Amsterdam, 1012 WX Amsterdam, Netherlands\\
\email{m.aliannejadi@uva.nl} \and
Elsevier Information Systems GmbH, Franklinstr. 63, 60486, Frankfurt am Main, Germany\\
\email{c.thorne.1@elsevier.com}
}

\maketitle

\begin{abstract}
Multi-label document classification is a traditional task in NLP. Compared to single-label classification, each document can be assigned multiple classes. This problem is crucially important in various domains, such as tagging scientific articles. Documents are often structured into several sections such as abstract and title. Current approaches treat different sections equally for multi-label classification. We argue that this is not a realistic assumption, leading to sub-optimal results. Instead, we propose a new method called \textbf{L}earning \textbf{S}ection \textbf{W}eights (LSW), leveraging the contribution of each distinct section for multi-label classification. Via multiple feed-forward layers, LSW learns to assign weights to each section of, and incorporate the weights in the prediction. We demonstrate our approach on scientific articles. Experimental results on public (arXiv) and private (Elsevier) datasets confirm the superiority of LSW, compared to state-of-the-art multi-label document classification methods. In particular, LSW achieves a 1.3\% improvement in terms of Macro F-1 while it achieves 1.3\% in terms of Macro recall on publicly available arXiv dataset.

\end{abstract}

\section{Introduction}

The structure of documents can vary from one domain and dataset to another, but typically, they tend to be divided into several \emph{sections}, viz., thematically connected spans of text, such as e.g., introduction, methods, results
and conclusion sections for experimental science articles. Furthermore, they can be ordered under potentially \emph{multiple} classes, giving rise to a multi-labelling classification task.

One might be tempted to ignore concatenate all the sections and apply a model \cite{10.1093/bioinformatics/btz142}. However, ignoring sections and assuming that they have equal importance is a weak assumption for document multi-label classification: in reality different sections contain different information, and do not contribute equally to the task, if at all. In our proposed approach, highlighting the important sections is performed through two layers of feed forward neural networks, which generate one weight per section. We additionally leverage these weights to explain and quantify the importance of each section. 

Our expectation is that the sections which include more substantial information, are assigned higher weights. However, the weights will change per document meaning that for one document, section A might be more important, while to classify another document, section B might be more important. Our work originates from the attention mechanism ~\cite{attention,10.1093/bioinformatics/btz142} where the goal is to assign higher weights to more important words. Our goal is to assign instead higher weights to more important sections (multi-label classification task). These learned weights can help researchers and content experts to better analyse the performance of the algorithm. 

The attention mechanism originates in neural machine translation~\cite{attention} where, given a sentence in a source language, the corresponding sentence in the target language is predicted. As opposed to traditional neural machine translation approaches~\cite{neural-machine} which treat words in the source sentence equally, attention-based models assign weights to each word in the source sentence to predict the next word in the target sentence. These weights can be interpreted as the importance of each word in the source language.

Different types of attention mechanisms have been proposed so far. Sparse Attention~\cite{sparse-attention,sparse2,sparse3} has been proposed to reduce the complexity and memory consumption of the original attention model. This type of attention focuses on a limited number of pairs for which attention weights should be computed, which yields a sparse matrix. Since there is a strong connection between matrix sparsity and graph neural networks~\cite{graph}, Graph Attention Networks~\cite{graphattention,graphattention2,graphattention3} have been proposed. These approaches, that try to maximize the utility of sparse matrices, suffer from a lack of strong theoretical background but can be used in specific models such as transformers~\cite{graphattention4} and generative approaches~\cite{graphattention5}. Performer~\cite{performer} was proposed to address these issues and decrease the run time of the attention-based models. Self-attention and multi-head attention~\cite{selfattention} were proposed to 1) capture the relationship between words in the source sentence and 2) obtain different possible relationships through multi-head attention.

All these approaches try to compute the attention weights at a word or sentence level ~\cite{sectionlevel}. As opposed to the approaches mentioned so far, we try to propose a new Learning Section Weights (LSW) approach. Our proposed LSW approach can measure the contribution of each section of the given article in the downstream task (e.g., classification). In this paper, the downstream task we have considered is multi-label text classification.

Indeed, as opposed to traditional BERT-based models~\cite{bert} which treat the text from different sections equally, we propose LSW network to learn the importance of different sections. The LSW network is trained jointly with the model used for classification, where back-propagation~\cite{bg}
will propagate the classification error through the LSW and classifier parameters.

Also, our proposed approach helps to have a deeper understanding of the different sections of a given article. Indeed, it brings additional information which helps data analysts and data scientists to draw conclusions based on the learned section weights.

\section{Methodology}\label{sec:methodology}

We focus our experiments on two datasets of scientific article snapshots,  containing a) abstracts, b) titles and c) keywords, and base our multi-label classifiers on the SciBERT \cite{beltagy-etal-2019-scibert} BERT model. As core baseline, we concatenate all sections and feed them to a BERT-base model to do classification. Thereafter we improve on this baseline by adding feed-forward layers on top of the BERT model. 

\begin{wrapfigure}{r}{0.5\textwidth}
\centering
\vspace*{-12mm}
\includegraphics[width=0.5\textwidth]{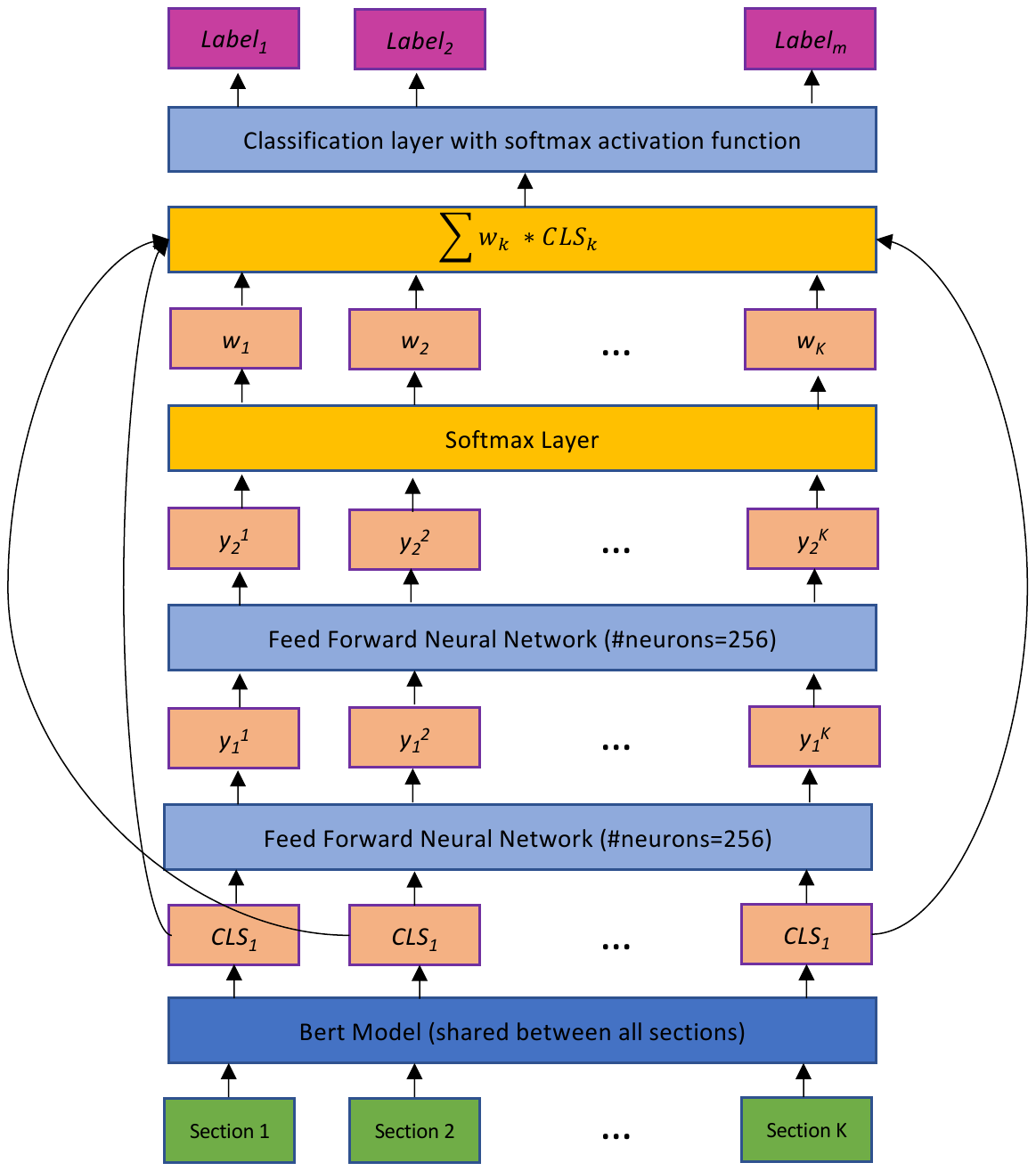}
\vspace*{-1.3cm}
\caption{LSW network architecture.}
\vspace*{-7mm}
\label{fig:LSW}
\end{wrapfigure}

The drawback with both approaches is that different sections are considered equally. To address this limitation, we propose to learn section weights to highlight more important sections and allow them to contribute to the multi-label classification task based on their importances.

Let $S=\{s_{k}\}_{k=1}^{K}$ represent all sections and $W=\{w_{k}\}_{k=1}^{K}$ represent the corresponding weights computed per section (representing sections importances). In the remainder, $\mathcal{X}$ will denote the set of documents to be classified, while $x \in \mathcal{X}$ represents a single document. $\text{CLS}_{k} \in \mathbb{R}^{d}$ ($d$ = 768) denotes the representation of section $k$ obtained from the BERT model for document $x$. $f_{\theta}$ : $\mathbb{R}^{d} \to \mathbb{R}^{p}$ ($p$ = 256) is a first linear layer (of parameters $\theta$). $g_{\eta}$ : $\mathbb{R}^{p} \to \mathbb{R}^{1}$ represents the second linear layer to compute section weights (of parameters $\eta$) which is also a feed-forward neural network. The output dimension of the second layer ($g_\eta$) equals to one as each section weight is a scalar representing the importance and contribution of the underlying section. 
Each layer is followed by a rectified linear unit ($\text{relu}$) activation followed by a (first) softmax layer to estimate section weights. Formally:
\begin{gather}
    y_{k}^{1} = \text{relu}(f_{\theta}(x)),\qquad
    y_{k}^{2} = \text{relu}(g_{\theta}(x)),\\
    w_{k} = \dfrac{\exp(y_{k}^{2})}{\sum_{k=1}^{K}{\exp(y_{k}^{2})}}.
    \label{eq:attention}
\end{gather}

After computing Eq. \ref{eq:attention}, the section weights can be used for weighting each section and then classifying the input document as follows: 
\begin{gather}
    y = \sum_{k=1}^{K}{w_k \cdot \text{CLS}_{k}},\qquad
    \text{output} = \text{softmax}(j_{\omega}(y))
\end{gather}
where
$y$ is the summarization of all sections (instead of simply concatenating them), and its dimension is equal to $\text{CLS}_{k}$ dimension (768) as $w_{k}$ is a scalar, while $j_{\omega}: \mathbb{R}^d \to \mathbb{R}^p \to \mathbb{R}^m$ ($m$ denotes the number of classes) represents a final stack of feed-forward linear layers (of parameters $\omega$) followed by a second softmax layer that performs the classification. Binary Cross-Entropy has been used as the loss function in our proposed LSW network. The proposed network has the following properties:
\begin{itemize}
    \item Since $\sum_{k=1}^{K}{w_K}$ = 1 (Eq. \ref{eq:attention}), further analysis per document classification can be done through section weights analysis.
    \item The contribution of each section in classification is determined by their corresponding section weights meaning that noisy (less useful) sections will contribute less to the multi-label classification. This improves classification results.
\end{itemize}
    
The architecture of the proposed LSW network is shown in Fig. \ref{fig:LSW}. Please note that in our proposed approach: 1) all sections share the same BERT model, and 2) the BERT model parameters are not frozen, meaning that its trainable parameters are updated through backpropagation.

\begin{table}[t]
\caption{Tuned hyperparameters used for training on our private (Elsevier) dataset.}\label{tab:hyper1}
\centering
\begin{tabular}{@{}|l|l|l|l|l|@{}}
\hline
Dataset &  Optimizer & Learning Rate  & \#Epochs & Minibatch\\
\hline
LSW & Adam & $e^{-5}$ & 5            & 32\\ 
Baseline \#1 &  Adam & $e^{-5}$ & 10 & 32\\ 
Baseline \#2 &  Adam & $e^{-5}$ & 5  & 32\\ 
Baseline \#3 &  Adam & $e^{-5}$ & 4  & 32\\ 
\hline
\end{tabular}
\label{table:results-hyper1}
\end{table}

\begin{table}[t]
\vspace*{-3mm}
\caption{Tuned hyperparameters used for training on arXiv datasets.}\label{tab:hyper2}
\centering
\begin{tabular}{@{}|l|l|l|l|l|@{}}
\hline
Dataset &  Optimizer & Learning Rate  & \#Epochs & Minibatch\\
\hline
LSW & Adam & $e^{-5}$ & 10           & 32\\ 
Baseline \#1 &  Adam & $e^{-5}$ & 10 & 32\\ 
Baseline \#2 &  Adam & $e^{-5}$ & 10 & 32\\ 
Baseline \#3 &  Adam & $e^{-5}$ & 10 & 32\\ 
\hline
\end{tabular}
\vspace*{-2mm}
\label{table:results-hyper2}
\end{table}

\begin{table}[t]
\caption{Datasets information.}\label{tab:dataset}
\centering
\scalebox{0.95}{
\begin{tabular}{@{}|l|l|l|l|@{}}
\hline
Dataset &  \#Documents & \#Classes  & Sections \\
\hline
Elsevier & 120,000 & 52 & [Abstract, Title, Keywords]\\ 
arXiv & 306,114 & 18 & [Abstract, Title]\\
\hline
\end{tabular}
}
\vspace*{-2mm}
\label{table:datasets}
\end{table}

\section{Experimental Setups and Results}\label{sec:results}

In this section, we discuss firstly baselines and how we fine-tuned the network then, and secondly the datasets used for our experiments. Finally, we show the results and a few section weights plots which illustrate respectively that our proposed LSW network can achieve both state-of-the-art results and bring explainability by measuring the contribution of each section.

\subsection{Baselines} \label{sec:baseline}

\myparagraph{Baseline \#1}: This architecture is quite similar to our proposed LSW network except for the section weights, which have been removed. Please note that in this case, BERT trainable parameters are not frozen, and they are updated due to classification errors (similar to our proposed LSW).

\begin{table}[t]
\caption{Multi-label classification results on arXiv dataset. Bold numbers represent the best performances.}\label{tab:results-ArXiv}
\centering
\resizebox{\textwidth}{!}{
\begin{tabular}{@{}|l|l|l|l|l|l|l|@{}}
\hline
Method &  Macro F-1 & Macro Precision & Macro Recall & Micro F-1 & Micro Precision & Micro Recall \\
\hline
LSW & \textbf{94.8}\%&  94.9\%& \textbf{95.0}\% & \textbf{96.5}\% & \textbf{96.4}\%& \textbf{96.6}\%\\
Baseline \#1 & 94.4\%& \textbf{95.3}\%& 93.7\%&  96.3\%&  96.2\%&  96.4\%\\
Baseline \#2 & 93.9\%& 94.4\%& 91.2\% & 94.5\% & 92.2\%& 92.3\%\\
Baseline \#3 & 89.1\%& 89.3\%& 89.2\% & 90.1\% & 90.0\%& 90.3\%\\
\hline
\end{tabular}
}
\vspace*{-2mm}
\label{table:results1}
\end{table}

\begin{table}[t]
\caption{Multi-label classification results on Elsevier dataset. Bold numbers represent the best performances.}\label{tab1}
\centering
\resizebox{\textwidth}{!}{
\begin{tabular}{@{}|l|l|l|l|l|l|l|@{}}
\hline
Method &  Macro F-1 & Macro Precision & Macro Recall & Micro F-1 & Micro Precision & Micro Recall \\
\hline
LSW & \textbf{58.0}\%& 69.2\% & \textbf{52.8}\% & \textbf{82.2}\% & \textbf{84.2}\%& \textbf{80.3}\%\\
Baseline \#1 & 56.7\%& \textbf{70.3}\%&52.7\% & 81.8\% & 83.8\%& 79.9\%\\
Baseline \#2 & 57.4\%& 67.1\%&\textbf{52.8}\% & 80.0\% & 81.0\%& 79.1\%\\
Baseline \#3 & 54.8\%& 66.0\%&51.5\% & 77.7\% & 78.3\%& 76.6\%\\
\hline
\end{tabular}
}
\vspace*{-2mm}
\label{table:results2}
\end{table}

\myparagraph{Baseline \#2}: This baseline concatenates all $\text{CLS}$ BERT representations and then applies a classification layer on top of these representations. Please note that BERT trainable parameters are not frozen, and they are updated at training time (similar to our proposed LSW).

\myparagraph{Baseline \#3}: The architecture of this baseline is identical to baseline \#1, with the difference that the BERT model parameters have now been frozen (i.e., simulates zero-shot learning).

\subsection{Dataset}

We have used both private (Elsevier) and public (arXiv) randomly sampled scientific article datasets to evaluate the performance of our proposed LSW%
\footnote{\url{https://tinyurl.com/y4su8pz2}} 
approach. Tab.\ref{tab:dataset} includes information regarding these two datasets. Note that each document contains an average of 200 words, giving rise to datasets comprising more than 20 million tokens each.

To fine-tune our proposed LSW network and the baselines mentioned above, we have used 10\% of our dataset as validation data. The hyperparameters that we have tuned are: 1) the choice of optimizer, 2) the learning rate, 3) batch-size and 4) the number of epochs. We report the best performing configurations in Tables \ref{table:results-hyper1}--\ref{table:results-hyper2}.

\begin{figure}[t]
\begin{subfigure}{.5\textwidth}\label{fig:sw1}
  \centering
  \includegraphics[width=.65\linewidth]{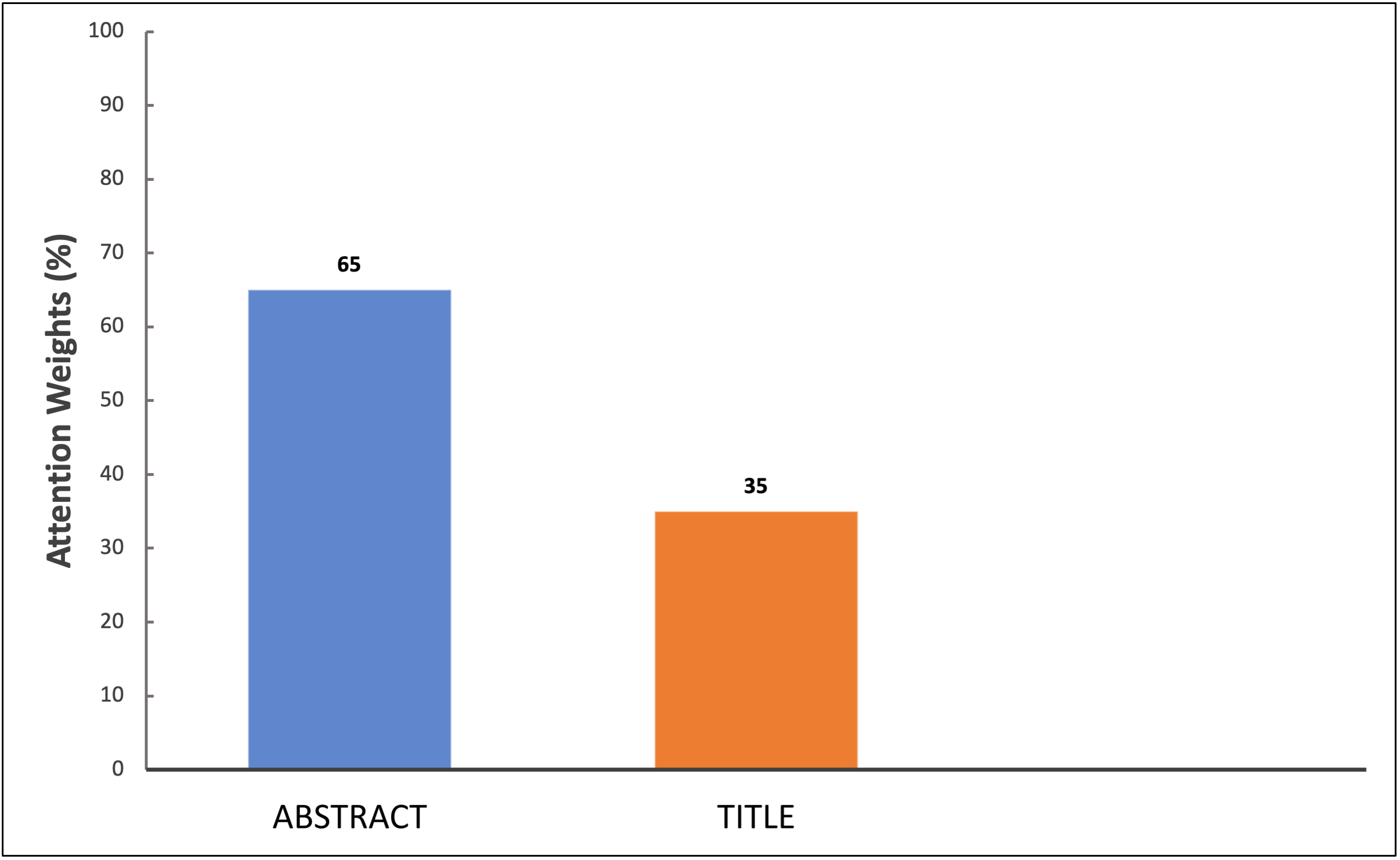}
\caption{}
  \label{fig:sfig2}
\end{subfigure}
\begin{subfigure}{.5\textwidth}\label{fig:sw2}
  \centering
  \includegraphics[width=.65\linewidth]{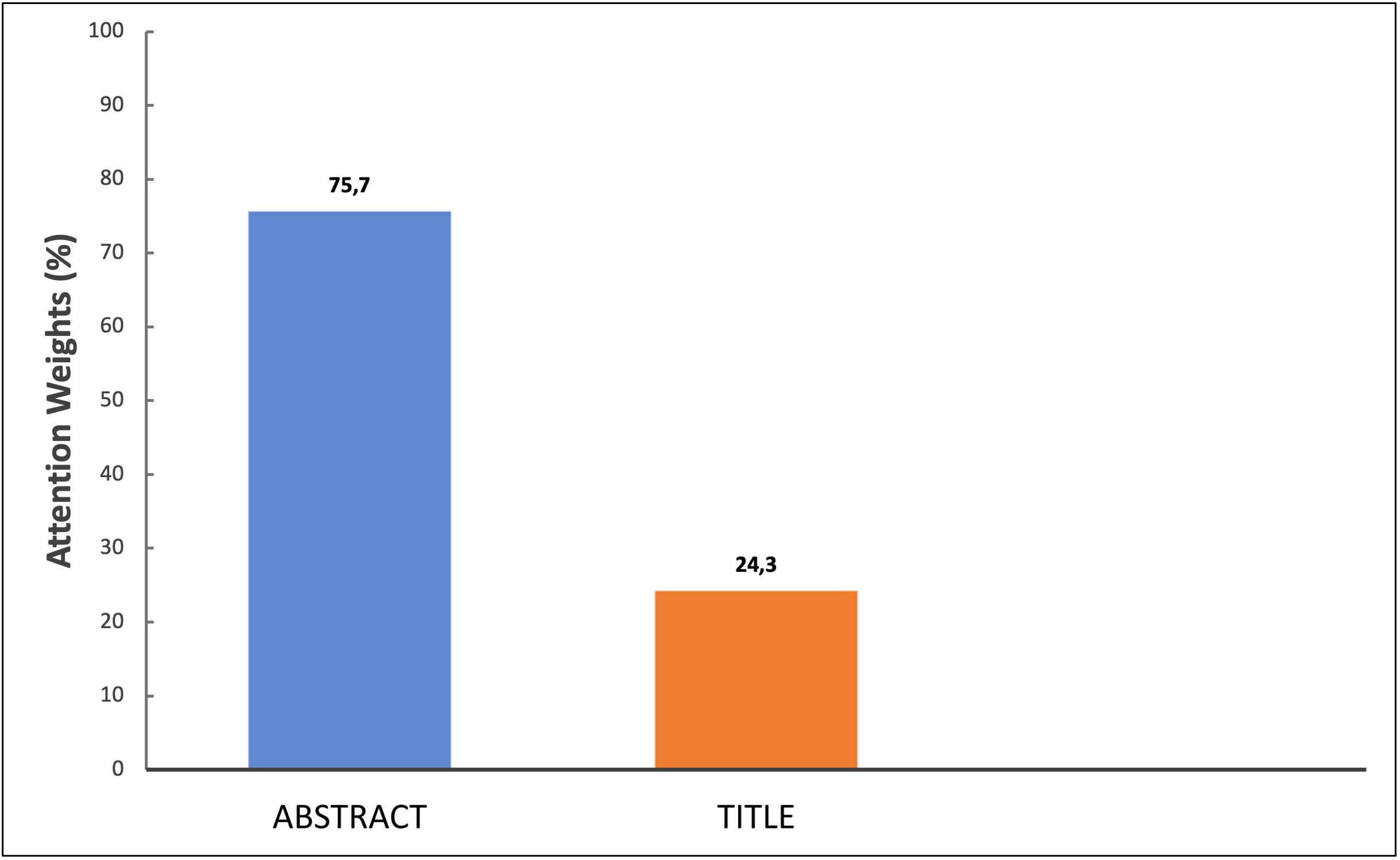}
\caption{}
  \label{fig:sfig1}
\end{subfigure}%
\centering
\begin{subfigure}{.5\textwidth}\label{fig:sw1}
 \centering
  \includegraphics[width=.65\linewidth]{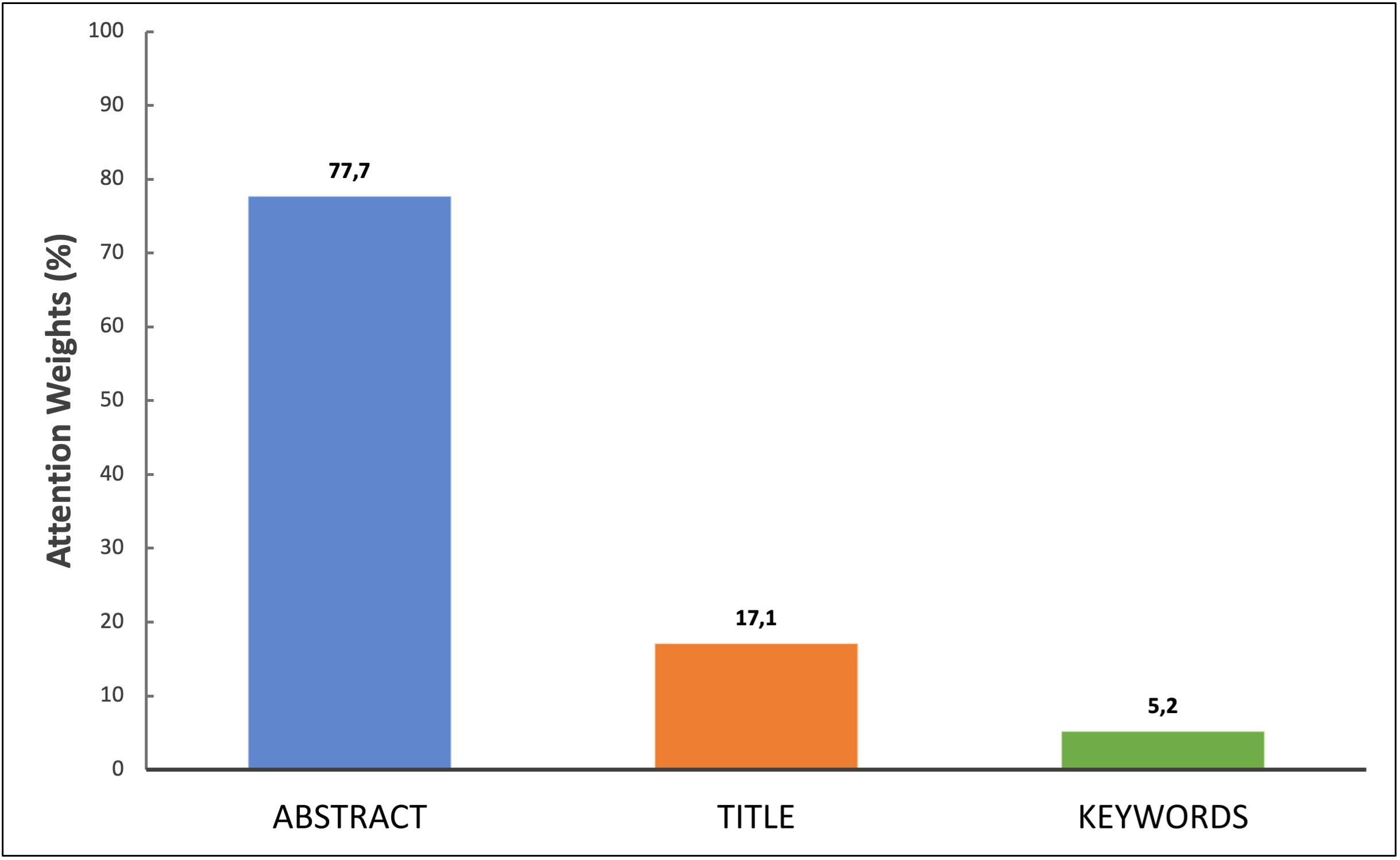}
\caption{}
  \label{fig:sfig2}
\end{subfigure}
\caption{Distribution of section weights indicating the importance of different sections. (a) shows section weights pertaining to a random document from arXiv dataset while (b) is related to section weights averaged on all arXiv test data. (c) shows section weights averaged on all Elsevier test data.}
\vspace*{-5mm}
\label{fig:att-plot}
\end{figure}

\subsection{Results} \label{sec:results}

We have evaluated the performance of our proposed LSW network and baselines mentioned in Sec. \ref{sec:baseline} using six different metrics, including Micro/Macro F1-score, Micro/Macro Precision and Micro/Macro Recall. The results shown in Tab. \ref{table:results1} and Tab. \ref{table:results2} indicate that our proposed algorithm can outperform baselines on the majority of the metrics. Indeed, we can notice that our proposed LSW approach is able to consistently outperform all baseline on all metrics except Macro Precision. The ability to surpass baselines on both arXiv (public) dataset and our private (Elsevier) dataset proves the stability of our proposed LSW approach meaning that LSW is able to handle real world case problems. Moreover, Fig. \ref{fig:att-plot} indicates that our proposed LSW network can assess the contribution of each section per document independently. As it can be noticed, per document, different sections contribute to the multi-label classification task differently, with the average section weights across all documents showing that, in general, the abstract plays the most important role in the tagging of the documents.

\section{Conclusions}\label{sec:conclusion}

In this paper, we  presented our proposed LSW network, which can assess the contribution of each section of an article in the multi-labelling classification downstream task. Indeed, this network can classify the underlying document and add explainability regarding the importance of each section. The section weights are updated and computed using gradient descent and backpropagation, which helps to obtain better results and utilise sections such that the classification error is minimised. The results of utilising this approach on both our Elsevier and arXiv datasets indicate that our proposed LSW network can achieve state-of-the-art results compared to different baselines. However, for future work, we will apply the proposed architecture to different tasks (e.g., clustering).

\section{Acknowledgements}
We extend our gratitude to the Elsevier Life Sciences Department, whose sponsorship and support made this research possible. We also extend our gratitude to all the many colleagues who contributed with their feedback to earlier drafts of this paper.

\bibliographystyle{splncs04}
\bibliography{sam}

\end{document}